\documentclass[journal]{IEEEtran}
\usepackage{amsmath}
\usepackage{fixmath}
\usepackage{float}
\usepackage{amssymb}
\usepackage{lipsum}
\usepackage{array,multirow,graphicx}
\usepackage[table]{xcolor}
\usepackage{subcaption}
\usepackage{lineno,hyperref}
\usepackage[ruled,longend]{algorithm2e}
\usepackage{booktabs}

\usepackage{hyperref}
\hypersetup{
	colorlinks,
	linkcolor=[rgb]{1.0, 0.0, 0.0},
	citecolor=[rgb]{1.0, 0.2, 0.2},
	urlcolor =[rgb]{0.0, 0.0, 0.8}
}
\ifCLASSINFOpdf
\else
\fi
\begin{document}
	\bstctlcite{IEEEexample:BSTcontrol}
	% to reduce the margin of equations and figure
	%\abovedisplayskip=0pt
	%\abovedisplayshortskip=0pt
	%\belowdisplayskip=0pt
	%%\belowdisplayshortskip=0pt
	%\abovecaptionskip=0pt
	%\belowcaptionskip=0pt
	
	\title{CoNIC: Colon Nuclei Identification and Counting Challenge 2022}
	

	\author{Simon~Graham$^1$, Mostafa Jahanifar$^1$, Quoc Dang Vu$^1$, Giorgos Hadjigeorghiou$^1$, Thomas Leech$^1$, David Snead$^2$, Shan E Ahmed Raza$^1$, Fayyaz Minhas$^1$, Nasir Rajpoot$^1$ \\ $^1$Department of Computer Science, University of Warwick, UK\\
	$^2$Department of Pathology, University Hospitals Coventry and Warwickshire, UK\\
{\tt\small conic2022@gmail.com}}

	\maketitle
	
	\begin{abstract}
% 		Nuclear segmentation, classification and quantification within Haematoxylin \& Eosin stained histology images is a fundamental prerequisite in the digital pathology work-flow, due to the ability for nuclear features to act as key diagnostic markers. The development of automated methods for nuclear segmentation and classification enables the quantitative analysis of tens of thousands of nuclei within a whole-slide pathology image, opening up possibilities of further analysis of large-scale nuclear morphometry. However, automated nuclear segmentation is faced with a major challenge in that there are several different types of nuclei, some of them exhibiting large intra-class variability such as the tumour cells. Additionally, some of the nuclei are often clustered together. To address these challenges, we present a novel convolutional neural network for automated nuclear segmentation that leverages the instance-rich information encoded within the vertical and horizontal distances of nuclear pixels to their centres of mass. These distances are then utilised to separate clustered nuclei, resulting in an accurate segmentation, particularly in areas with overlapping instances. We demonstrate state-of-the-art performance compared to other methods on four independent multi-tissue histology image datasets. Furthermore, we propose an interpretable and reliable evaluation framework that effectively quantifies nuclear segmentation performance and overcomes the limitations of existing performance measures.
		Nuclear segmentation, classification and quantification within Haematoxylin \& Eosin stained histology images enables the extraction of interpretable cell-based features that can be used in downstream explainable models in computational pathology (CPath). However, automatic recognition of different nuclei is faced with a major challenge in that there are several different types of nuclei, some of them exhibiting large intra-class variability. To help drive forward research and innovation for automatic nuclei recognition in CPath, we organise the Colon Nuclei Identification and Counting (CoNIC) Challenge. The challenge encourages researchers to develop algorithms that perform segmentation, classification and counting of nuclei within the current largest known publicly available nuclei-level dataset in CPath \cite{graham2021lizard}, containing around half a million labelled nuclei. Therefore, the CoNIC challenge utilises over 10 times the number of nuclei as the previous largest challenge dataset for nuclei recognition \cite{verma2021monusac2020}. It is important for algorithms to be robust to input variation if we wish to deploy them in a clinical setting. Therefore, as part of this challenge we will also test the sensitivity of each submitted algorithm to certain input variations.
	\end{abstract}
	

	% \fi
	\IEEEpeerreviewmaketitle
	
	\section{Introduction} \label{section:intro}
	\IEEEPARstart{D}EEP learning models have revolutionised the field of computational pathology (CPath), partly due to their ability in leveraging the huge amount of image data contained in multi-gigapixel whole-slide images (WSIs). Yet, utilising CNNs in an end-to-end manner for slide-level prediction can lead to poor explainability, due to a high-level of model complexity with limited feature interpretability \cite{tosun2020explainable, holzinger2017towards, jaume2020quantifying}. Explainable AI in CPath may be preferable because it can ensure algorithmic fairness, identify potential bias in the training data, and ensure that the algorithms perform as expected \cite{gilpin2018explaining}. 
	
	In order to extract meaningful human-interpretable features from the tissue, accurate \textbf{localisation} of clinically relevant structures is often an important initial step. For example, features indicative of nuclear morphology first require each nucleus to be segmented and can then be directly used in downstream tasks, such as predicting the cancer grade \cite{alsubaie2018bottom} and survival analysis \cite{lu2020prognostic}. Sometimes, rather than exploring features indicative of the nuclear shape, it may be of interest to accurately \textbf{quantify} different types of cells, which may negate the need to explicitly localise each nuclear boundary. For example, the counts of tumour cells and lymphocytes have recently been used as a powerful prognostic marker \cite{fridman2012immune}. 
	
	We must also ensure that developed algorithms are not sensitive to input variation if we wish for them to be succesfully used in the clinic. For example, recent work showed that adding a subtle change could consistently change the prediction made by a computational pathology system. \cite{foote2021now}.
	
	As a result of the above, we organise the \underline{\textbf{Co}}lon \underline{\textbf{N}}uclei \underline{\textbf{I}}dentification and \underline{\textbf{C}}ounting (CoNIC) challenge. We hope that the challenge will drive forward research and innovation for automated nuclear localisation, classification and quantification methods, enabling the extraction of interpretable cell-based features for downstream applications in CPath. In the challenge we utilise around half a million labelled nuclei, which is over 10 times the number of nuclei as the previous largest challenge dataset for nuclei recognition \cite{verma2021monusac2020}. In order to assess each algorithm's robustness, we will also analyse the ability of each model to handle input variation during evaluation. 
	
	 We provide all relevant information, including details on the tasks, data and evaluation on our webpage\footnote{https://conic-challenge.grand-challenge.org}. We encourage participants to regularly check this for challenge updates. 
	 
	 \begin{figure*}[t]
		\centering
        \includegraphics[width=0.9\textwidth]{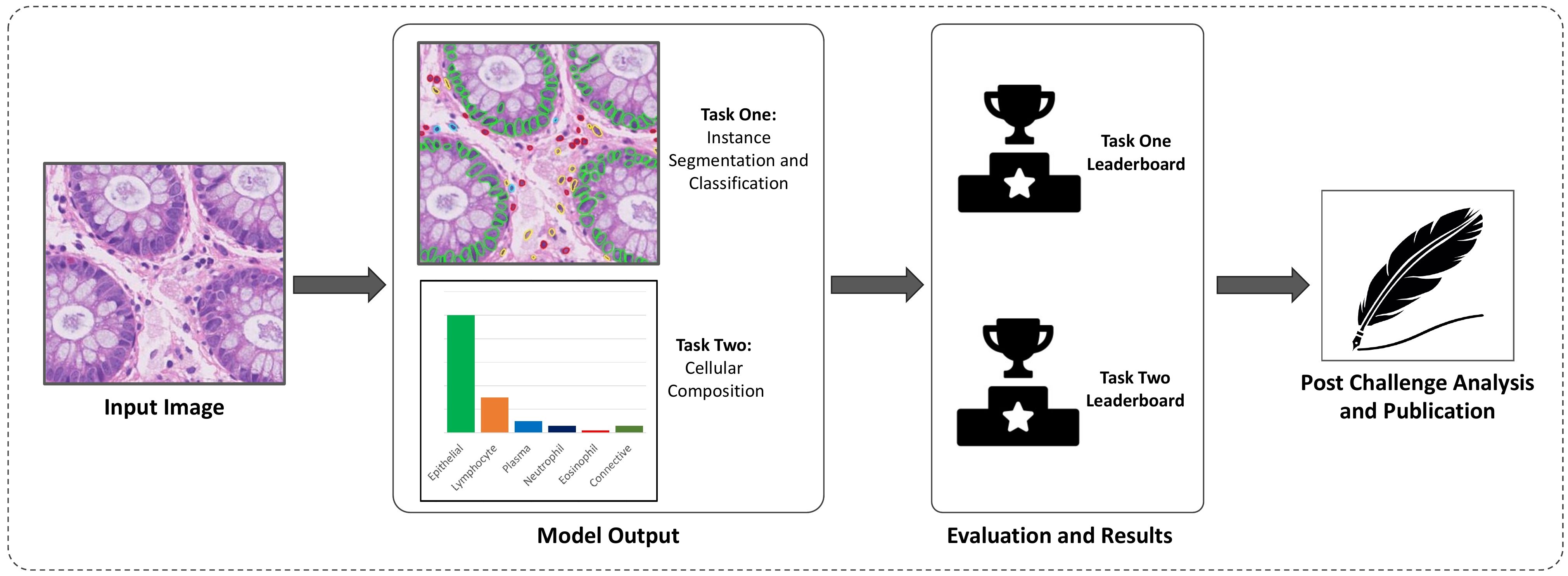}
		\caption{Overview of the tasks considered in the CoNIC challenge. Each task is considered independently, but a single model can perform multiple tasks. For example, a model that performs instance segmentation and classification can perform both tasks. A participant does \textbf{not} need to perform all tasks in the challenge.} 
		\label{fig:overview}
	\end{figure*}

    \section{CoNIC Challenge Details}
    \label{section:challenge}
    \subsection{Tasks}
    
    In this challenge, we encourage participants to complete 2 tasks:
    
    \begin{itemize}
        \item Nuclear instance segmentation and classification
        \item Cellular composition (counting)
    \end{itemize}
    
    The type of algorithm developed will determine what tasks it can be applied to. For example, a model developed for nuclear instance segmentation and classification can perform both tasks, whereas a model built for predicting cellular composition will not be able to naturally localise nuclear boundaries.
    
    A participant does not need complete \textbf{both} tasks, but may focus on a single task such as cellular composition. For example, this has been recently done by Dawood \textit{et al.} \cite{dawood2021albrt} where the counts of different cell types were predicted without explicitly localising each nucleus. An overview of the tasks considered during the challenge is shown in Fig. \ref{fig:overview}.
    % As a result, we consider splitting the challenge into three tasks where they can be attacked separately (such as direct cell counting) or sequentially (detection and then classification for cell counting).
    
    A thorough analysis on the robustness of each method to input variation will be performed \textbf{after} the challenge and described in the post-challenge publication. Therefore, despite it not being used to rank the leaderboard, participants should consider this during model development. Further information on this will be described during the challenge.

    \subsection{Challenge Data}
    \label{section:data}
    \subsubsection{Lizard dataset}
    \label{section:lizard}
    In the CoNIC challenge, we utilise our developed Lizard dataset \cite{graham2021lizard}, which consists of histology image regions of colon tissue from six different dataset sources at 20$\times$ objective magnification, with full segmentation annotation for different types of nuclei. In particular we provide the nuclear class label for epithelial cells, connective tissue cells, lymphocytes, plasma cells, neutrophils and eosinophils and therefore models trained on this data may be used to help effectively profile the colonic tumour micro-environment. We choose to focus on nuclei from colon tissue to ensure that our dataset contains images from a wide variety of different normal, inflammatory, dysplastic and cancerous conditions in the colon - therefore increasing the likelihood of generalisation to unseen examples. In the original dataset, the images are stored as .png files, whereas the labels are stored as .mat files, containing the instance map, nuclear categories, bounding boxes and nuclei counts. 
    
\begin{figure*}[t]
		\centering
        \includegraphics[width=0.95\textwidth]{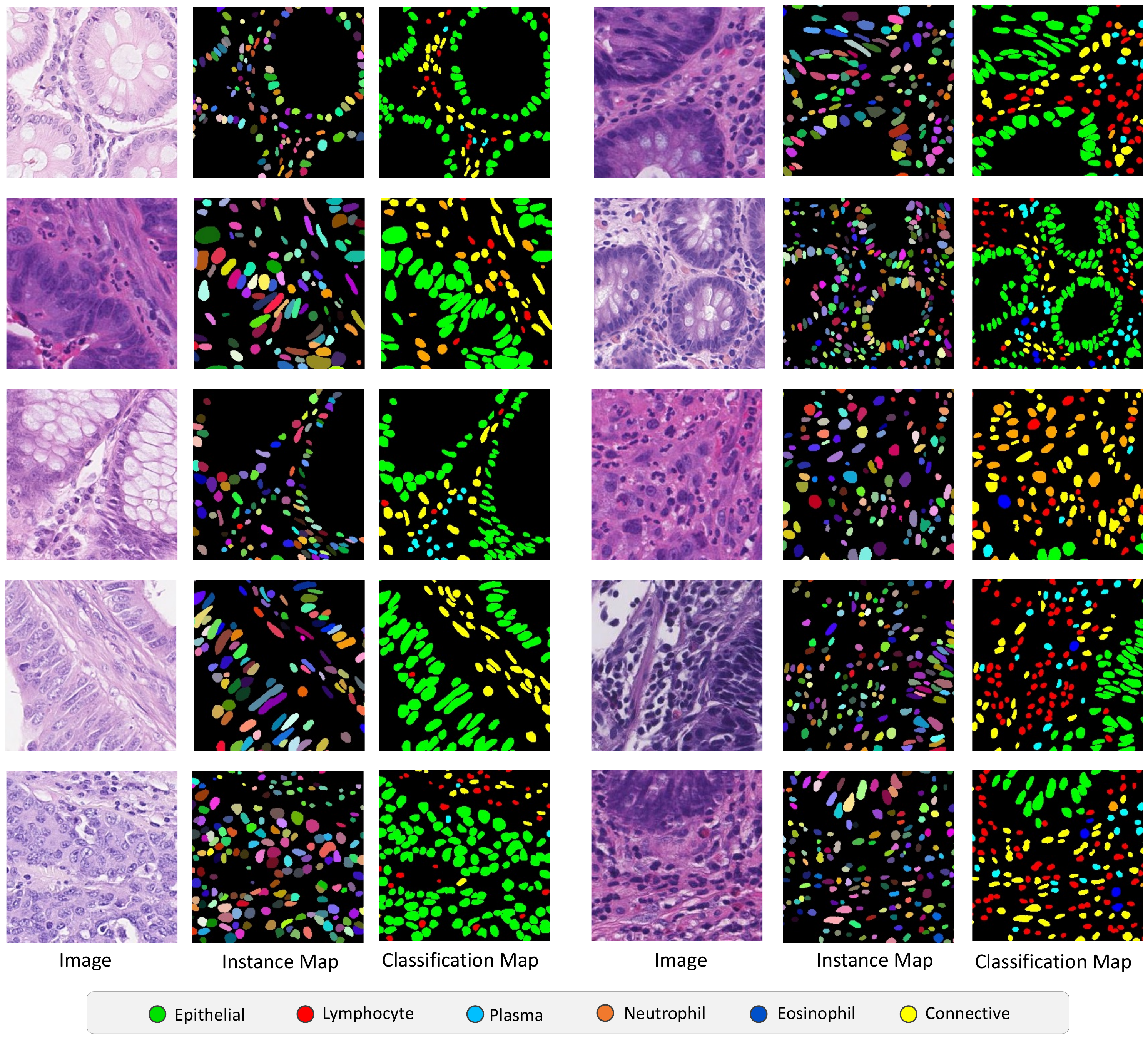}
		\caption{Example patches along with the respective ground truth for segmentation and classification. Different colours within the instance map denote separate nuclei. The legend at the bottom of the figure indicate the meaning of the colours within the classification map.} 
		\label{fig:patches}
	\end{figure*}

\begin{figure*}[t]
		\centering
        \includegraphics[width=0.97\textwidth]{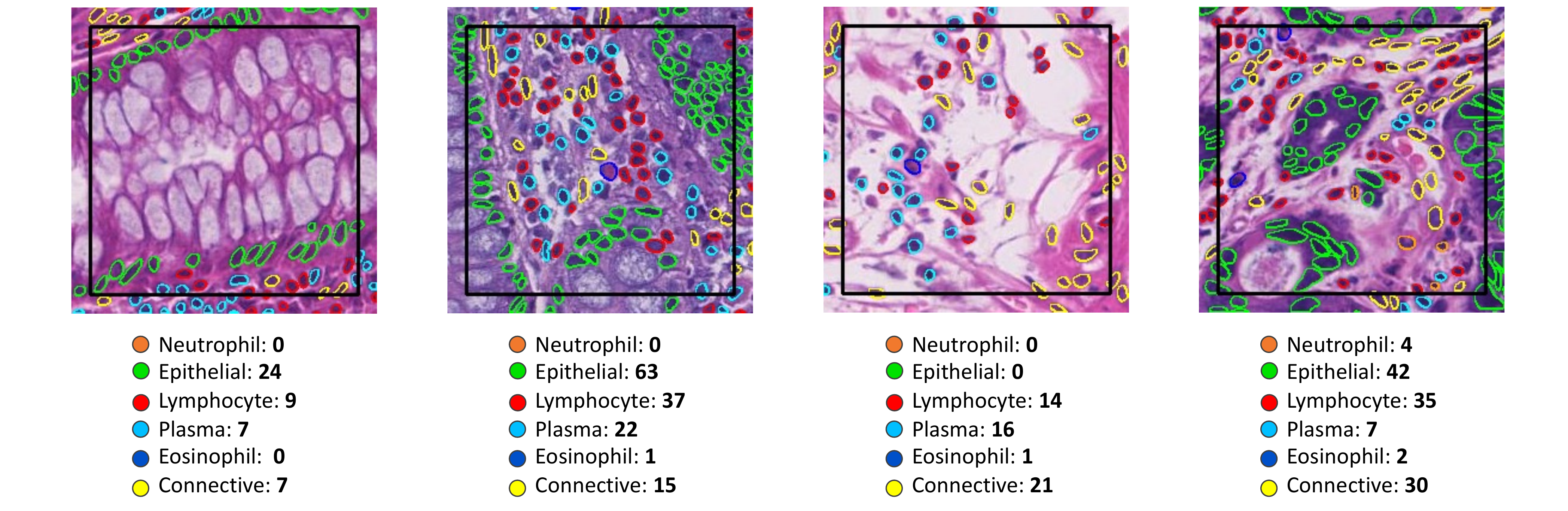}
		\caption{Example patches along with the respective counts per nuclear type. The size of all patches are size 256$\times$256, but the counting is only performed in the central 224$\times$224 region, shown by the black box.} 
		\label{fig:counting}
	\end{figure*}

     For use within the challenge, we provide 4,981 patches of size 256$\times$256  extracted from the original Lizard dataset. The download link for the patch-level dataset can be found on the challenge website. Each RGB image patch is associated with an instance segmentation map and a classification map, which are both also of size 256$\times$256. The instance segmentation map uniquely labels each nucleus by containing values ranging from 0 (background) to $N$ (number of nuclei). The classification map provides the class for each pixel within the patch. Specifically, the map contains values ranging from 0 (background) to $C$ (number of classes). The RGB images and segmentation/classification maps are each stored as a single numpy array. The RGB image array has dimensions 4981$\times$256$\times$256$\times$3 and is of type unsigned 8-bit integer, whereas the segmentation \& classification map array has dimensions 4981$\times$256$\times$256$\times$2 and  is of type unsigned 16-bit integer. Here, the first channel is the instance segmentation map and the second channel is the classification map. Examples of patches along with their respective instance segmentation and classification maps are shown in Figure \ref{fig:patches}. For the nuclei counts, we provide a single .csv file, where each row corresponds to a given patch and the columns determine the counts for each type of nucleus. The row ordering is in line with the order of patches within the numpy files. For this, a given nucleus is considered present in the image if any part of it is within the central 224$\times$224 region within the patch. This ensures that a nucleus is only considered for counting if it lies completely within the original 256$\times$256 image. We show example image patches along with the corresponding counts for each nuclear type in Fig. \ref{fig:counting}.
     
    All data is associated with a \textbf{non-commercial} creative commons license and can be downloaded from our challenge webpage\footnote{https://conic-challenge.grand-challenge.org/Data}.
    
    \subsubsection{Training Data}
    For training, we release the Lizard data from the following dataset sources: DigestPath, CRAG \cite{graham2019mild}, GlaS \cite{sirinukunwattana2017gland}, CoNSeP \cite{graham2019hover} and PanNuke \cite{gamper2020pannuke}, consisting of a total of 431,913 labelled nuclei. Of these nuclei, 210,372 are epithelial, 92,238 are lymphocytes, 24,861 are plasma cells, 4,116 are neutrophils and 2,979 are eosinophils. Therefore, our released patch-level dataset also contains data from the same described data sources. Note, despite using non-overlapping patches, there are more nuclei in the patch-level dataset because we resample patches at the image boundaries in the case that the image width or height is not exactly divisible by the patch size.
    
    \subsubsection{Test Data}
    For the CoNUIC challenge, there will be a preliminary test phase that can be used to test initially developed algorithms before conducting the final test phase. Note, test images will not be directly sent to participants, due to using code-based submissions in this challenge. Similar to our the patch dataset released for training, all data at test time will contain patches of size 256$\times$256. Therefore, it should be ensured that models trained by participants use patches of the same size. Note, if participants want to extract overlapping patches from the original dataset, then this is allowed. To make this easy, we provide code that enables patch extraction with a user-defined overlap on our GitHub page\footnote{https://github.com/TissueImageAnalytics/CoNIC}. 
    
    The preliminary test set will will consist of images taken from the TCGA database, containing around 25K labelled nuclei. Later in the challenge, we will utilise the entire test set, which will be used to quantify model performance and to determine the final ranking. The final test set will consist of 63,266 labelled nuclei in images taken from TCGA (including those provided in the preliminary test set), as well as around 50-100K labelled nuclei from an internal colon dataset. No images from TCGA will be used during model training and therefore is a completely external test set. Also, because TCGA is a multi-centre dataset, we will also ensure that the final test set contains TCGA centres that were not seen in the preliminary test set. 
    
    \section{Submission and Evaluation}
    % \subsection{Submission format}
    % Individuals are required to register for the challenge on the dedicated website in order to be able to download the train/test data and participate in the challenge. Challenge participants are asked to download the test data, apply their method on it and submit the results in a particular format on the website. Submissions for the final test phase should be accompanied with a short technical paper describing their methods. 
    % More information on the desired output format, submission instructions, and technical paper write up will be released later with the launch of the challenge website. 
    
    \subsection{Evalutation Website}\label{section:evaluation}
    We use the Grand Challenge platform to host the CoNIC challenge, which enables participants to seamlessly register and submit results. On the website, we provide detailed information describing the tasks, data and evaluation metrics. Submission instructions will also be provided on the website at a later date. Our page also contains a forum that can be used by participants to ask any challenge-related questions.
    
    \subsection{Making a Submission}
    Submissions will be made by submitting code as a Docker container. Therefore, participants will not have access to the test images, which ensures that the test set remains completely unseen for evaluation purposes. Further instructions on Docker based submissions, along with detailed examples, will be given at a later date. In addition to submitting the code, participants must also provide a manuscript describing their method. Below we provide further details on the requirements for submission.
    
    \subsection{Model output format}
    \subsubsection{Instance segmentation and classification} For this task, participants should return results as a single Numpy array in the same format as the patch-level training set. In particular, the submitted array should be of size $n\times$256$\times$256$\times$2, where $n$ is the number of test patches. Like the training set, the first channel should be a 2-dimensional instance map prediction of type unsigned 16-bit integer (uint16) containing pixel values from 0 to $N$, where 0 is background and $N$ is the number of predicted nuclei. All pixels belonging to the same nucleus should be given the same pixel value. The second channel should be the classification map, which again should be 2-dimensional and of type uint16. Here, values should be between 0 and $C$, where 0 is background and $C$ is the number of classes.
    
    \subsubsection{Cellular composition} We ask participants to generate a single ``csv'' file for the task of predicting cellular composition. Each row should indicate a different input patch and the columns should be named as follows: \textit{epithelial}, \textit{lymphocyte}, \textit{plasma}, \textit{neutrophil}, \textit{eosinophil} and \textit{connective}. Each column corresponding to a certain nuclear type should contain the corresponding count predictions. Here, the counts do not necessarily need to be integers.
    
    \subsection{Evaluation Metrics}
    \label{section:eval}
    Model evaluation will be done automatically using the Grand Challenge platform, but we will also provide evaluation code on the challenge GitHub page to enable participants to enable thorough experimentation with our utilised metrics. We will use one metric per task, which will determine the final standings of both leaderboards. For the metrics, we compute the statistics independently per class and the results are then averaged, which ensures there is no bias towards classes with a greater number of examples.
    \subsubsection{Nuclear instance segmentation and classification}
    For this task, we will use multi-class panoptic quality  \cite{kirilov2018panoptic_quality} ($PQ$) to evaluate the overall instance segmentation and classification performance. For each type $t$, we define $PQ$ as:

	\begin{equation}
	\small
	\mathcal{PQ}_t= 
	\underbrace{\frac{|TP_t|}{|TP_t|+\frac{1}{2}|FP_t|+\frac{1}{2}|FN_t|}}_{\text{Detection Quality(DQ)}}
	\times
	\underbrace{{\frac{\sum_{(x_t,y_t)\in{TP}}{IoU(x_t,y_t)}}{|TP_t|}}}_{\text{Segmentation Quality(SQ)}}
	\end{equation}

    \noindent where \textit{x} denotes a ground truth GT (GT) instance, \textit{y} denotes a predicted instance, and IoU denotes intersection over union. Setting IoU(\textit{x,y})$>$0.5 will uniquely match \textit{x} and \textit{y}. This unique matching therefore splits all available instances of type $t$ within the dataset into matched pairs (TP), unmatched GT instances (FN) and unmatched predicted instances (FP). Henceforth, we define multi-class $PQ$ ($mPQ^+$) as the task ranking metric, which is the average per-class $PQ$. Specifically, this is defined as:
	
	\begin{equation}
	\small
	m\mathcal{PQ^+}=\frac{1}{T}\sum_{t}{PQ_t}
	\end{equation}
	
	\noindent
	where $T$ is the number of types considered within the dataset. Specifically, within this challenge, $T$ is 6. Note, for $mPQ^+$ we calculate the statistics over \textbf{all} images to ensure there are no issues when a particular class is not present in a given patch. This is different to $mPQ$ calculation used in previous publications, such as PanNuke \cite{gamper2020pannuke}, MoNuSAC \cite{verma2021monusac2020} and in the original Lizard paper \cite{graham2021lizard}, where the $PQ$ is calculated for each image and for each class before the average is taken.

    \subsubsection{Cellular composition}
    
    For assessing the regression count, we use the multi-class coefficient of determination ($R^2$) to determine the correlation between the predicted and true counts. For this, the statistic is calculated for each class independently and then results are then averaged. In particular, for each nuclear category $t$ the correlation of determination is defined as follows:

	\begin{equation}
	\small
	\mathcal{R}^2_t=\frac{\sum_{i}{(m_i - k_i)^2}}{\sum_{i}{(m_i - \overline{k})^2}}
	\end{equation}
    
    \noindent where  $m$ is the GT counts for type $t$ and $k$ is its predicted counts. Here, $\overline{k}$ is the mean of all predicted counts. Here, the numerator is commonly referred to as the sum of squares of residuals and the denominator as the total sum of squares. The numerator and denominator are computed for all examples $i$ existing in class $t$.
    
    \subsection{Manuscript Submission Format}
    We require participants to submit a manuscript detailing their method along with the submission of their code. This enables us to understand what kind of models were successful and it also allows us to make sure that different submissions from the same research group are different. Submitted manuscripts will also help the organisers populate the necessary information when completing the post-challenge publication.
    
    Specifically, participants must submit a paper to ArXiv in Lecture Notes Computer Science (LNCS) format, with a maximum of 4 pages (excluding references). The manuscript should detail: data split, data pre-processing, model design, post processing and any other necessary details that were used in the challenge.

    \section{Baseline}\label{section:baseline}
    
    To help guide researchers on the ability of their developed model to recognise various nuclei, we shall present a baseline result as part of this challenge. For this, we will use HoVer-Net \cite{graham2019hover}, which is a top-performing model for nuclear segmentation and classification within the field of computational pathology. HoVer-Net outputs an instance segmentation mask, along with prediction of each nuclear category. These results can therefore be leveraged to directly determine the cellular composition. We will provide this baseline in the form of an example notebook on the challenge GitHub page, which can be interactively used by the participants. Further information on the data split and training details used for the baseline will also be given in the respective notebook.
    
    \section{Organisation}
    \subsection{Timeline}
    \begin{itemize}
        \item \textbf{November 20th 2021}: Release of training data and evaluation code.
        \item \textbf{December 13th 2021}: Release of baseline method.
        \item \textbf{February 13th 2022}: Preliminary test set submission opens.
        \item \textbf{February 27th 2022}: Preliminary test set submission closes and final test set submission opens.
        \item \textbf{March 6th 2022}: Final test set submission closes.
    \end{itemize}
    
    \subsection{Participation Rules \& Policies}
    There several policies that we ask participants to carefully read for participating in this competition:
    \begin{itemize}
        \item All final submissions should be in the form of Docker containers and follow the provided template provided by the organisers to be tested successfully.
        \item It is the user's responsibility to check the sanity of  their algorithm's Docker container during the preliminary test period before submitting it for the final test set.
        \item Only fully automated methods will be accepted, as the submission is in the form of Docker containers. It is not possible to submit manual annotations or interactive methods.
        \item Submissions will be ranked solely based on the performance on the final test set and according to the criteria explained in Section \ref{section:eval}.
        \item Using pre-trained models on general purpose datasets (like ImageNet) is allowed. 
        \item Use of external pathology data for model development is \textbf{not} permitted.
        \item Writing a technical paper for final submission is compulsory, otherwise the submission will not be considered.
        \item Users/teams can have multiple submissions on the preliminary test set to verify the quality of their methods. However, the number of submissions per day per user and for each task is limited to 1 in order to restrict test data hacking. 
        \item For the final test set, only 1 submission is allowed for each user/team.
        \item Users can form a team and participate together. Teams can have different submissions as long as their submissions vary considerably based on the described methodology in their technical papers. Otherwise, organisers have the right to disqualify those submissions at their discretion.
        \item Members from the organising team cannot participate in this challenge. If there is a baseline method in the leaderboard submitted by the organisers, that method will not be taken into account for the final ranking.
    \end{itemize}

    \subsection{Post-Challenge Publication}
    After the challenge, we will perform a significant analysis of the results and in a in a peer-reviewed journal. For this, we will investigate the performance of models on additional data and potentially other tissues to test each model's generalisation capability. We will also perform an analysis on each model's robustness to certain input variations. This is particularly important if we wish the deploy these models in a clinical setting.
	 	 
	\subsection{Sponsorship and Prizes}\label{section:sponsorship_and_prizes}
	For the CoNIC challenge, we will get one sponsor for each of the tasks, which will be awarded to winners of the challenge. Further information on this will be provided at a later date.

	\ifCLASSOPTIONcaptionsoff
	\newpage
	\fi

	
	
	%\bibliographystyle{IEEEtran}
	\bibliographystyle{IEEEtran}
	\bibliography{IEEEref.bib}

\end{document}